\documentclass{article}

    \PassOptionsToPackage{numbers, compress}{natbib}

     \usepackage[final]{neurips_2019}

\usepackage[utf8]{inputenc} 
\usepackage[T1]{fontenc}    
\usepackage{hyperref}       
\usepackage{url}            
\usepackage{booktabs}       
\usepackage{amsfonts}       
\usepackage{nicefrac}       
\usepackage{microtype}      

\usepackage[acronym,smallcaps,nowarn,section,nogroupskip,nonumberlist]{glossaries}

\usepackage[font=small,skip=0pt]{caption}

\usepackage{subcaption}
\usepackage[linesnumbered,ruled,vlined]{algorithm2e}
\usepackage{bm} 
\usepackage{enumitem} 

\usepackage{mlmacros} 
\usepackage[nameinlink,capitalise]{cleveref} 

\usepackage{titlesec}

\titlespacing*{\section}{0pt}{.1\baselineskip}{.1\baselineskip}
\titlespacing*{\subsection}{0pt}{0.1\baselineskip}{0.1\baselineskip}

\variables{l,e,s,x,y,z,b,k,g,y,n,w,D,P,S}

\probdists{p,q}
\probdists[pd]{p^D}
\probdists[pp]{p^P}
\probdists[qd]{q^D}
\probdists[qp]{q^P}

\title{R-SQAIR:Relational Sequential Attend, Infer, Repeat}

\author{%
  Aleksandar Stani{\'c}
    \\
  Swiss AI Lab, IDSIA, USI, SUPSI \\
  Lugano, Switzerland \\
  \texttt{aleksandar@idsia.ch} \\
   \And
   J\"{u}rgen Schmidhuber \\
  Swiss AI Lab, IDSIA, USI, SUPSI, NNAISENSE \\
  Lugano, Switzerland \\
  \texttt{juergen@idsia.ch} \\
}

\begin{document}

\maketitle

\begin{abstract}
Traditional sequential multi-object attention models rely on a recurrent mechanism to infer object relations. 
We propose a relational extension (R-SQAIR) of one such attention model (SQAIR) by endowing it with a module with strong relational inductive bias that computes in parallel pairwise interactions between inferred objects. 
Two recently proposed relational modules are studied on tasks of unsupervised learning from videos. 
We demonstrate gains over sequential relational mechanisms, also in terms of combinatorial generalization.
\end{abstract}

\section{Introduction}

Numerous studies \citep{xu1996infants, kellman1983perception, spelke1995spatiotemporal, baillargeon1985object, saxe2006perception} show that infants quickly develop an understanding of intuitive physics, objects and relations in an unsupervised manner.
To facilitate the solution of real-world problems, intelligent agents should be able to acquire such knowledge \cite{van2019perspective}.
However, artificial neural networks are still far from human-level understanding of intuitive physics.

Existing approaches to unsupervised learning about objects and relations from visual data can be categorized into either parallel \cite{greff2016tagger, greff2017neural, greff2019multi} or sequential \cite{schmidhuber91artificial, schmidhuber93ratioicann, eslami2016attend,  kosiorek2018sequential, crawford2019spatially, burgess2019monet, yuan2019generative}, depending on the core mechanism responsible for inferring object representations from a single image. 
One model from the former group is Tagger \cite{greff2016tagger} which applies the Ladder Network \cite{rasmus2015semi} to perform perceptual grouping.
RTagger \cite{premont2017recurrent} replaces the Ladder Network by a Recurrent Ladder Network, thus extending Tagger to  sequential settings.
NEM \cite{greff2017neural} learns object representations using a spatial mixture model and its relational version R-NEM \cite{van2018relational} endows it with a parallel relational mechanism.
The recently proposed IODINE \cite{greff2019multi} iteratively refines inferred objects and handles multi-modal inputs.

On the other hand, the sequential attention model AIR \cite{eslami2016attend} learns to infer one object per iteration over a given image. 
Contrary to NEM, it extracts object glimpses through a hard attention mechanism \cite{schmidhuber91artificial} and processes only the corresponding glimpse.
Furthermore, it builds a probabilistic representation of the scene to model uncertainty. 
Many recent models have AIR as the core mechanism: SQAIR \cite{kosiorek2018sequential} extends AIR to sequential settings, similarly does DDPAE \cite{hsieh2018learning}. 
SPAIR \cite{crawford2019spatially} scales AIR to scenarios with many objects and SuPAIR\cite{stelzner2019faster} improves speed and robustness of learning in AIR.
The recent MoNET \cite{burgess2019monet} also uses a VAE and a recurrent neural network (RNN) to decompose scenes into multiple objects.
These methods usually model relations by a sequential relational mechanism such as an RNN which limits their relational reasoning capabilities\cite{battaglia2018relational}.

Here we present Relational Sequential Attend, Infer, Repeat  (R-SQAIR) to learn a generative model of intuitive physics from video data.
R-SQAIR builds on SQAIR which we augment by a mechanism that has a strong relational inductive bias \citep{battaglia2016interaction, van2018relational, santoro2018relational}. 
Our explicit parallel model of pairwise relations between objects is conceptually simpler than a sequential RNN-based model that keeps previous interactions in its memory and cannot directly model the effects of interactions of previously considered objects. 
Our experiments demonstrate improved generalization performance of trained models in new environments.


\section{Relational Sequential Attend Infer Repeat}

\textbf{Attend, Infer, Repeat (AIR)} \citep{eslami2016attend} is a generative model that explicitly reasons about objects in a scene. 
It frames the problem of representing the scene as probabilistic inference in a structured VAE. 
At the core of the model is an RNN that processes objects one at a time and infers latent variables $\bz = \{\bz_\mathrm{what}^{\mathrm{i}}, \bz_\mathrm{where}^{\mathrm{i}}, z_\mathrm{pres}^{\mathrm{i}}\}_{i=1}^n$, where $n\in\mathbb{N}$ is the number of objects. 
The continuous latent variable $\bz^{\mathrm{what}}$ encodes the appearance of the object in the scene and $\bz^{\mathrm{where}}$ encodes the coordinates according to which the object glimpse is scaled and shifted by a Spatial Transformer \citep{jaderberg2015spatial}. 
Given an image $\bx$, the generative model of AIR is defined as follows:
\vspace{-5pt}
\begin{align}
\label{eq:air}
p_\theta(\bx) = \sum\limits_{n=1}^N p_\theta (n) \int p_\theta^z (\bz | n) p_\theta^x(x|\bz) d\bz,
\end{align}
where $p_\theta (n) = \mathrm{Geom} (n \mid \theta)$ represents the number of objects present in the scene, $p_\theta^z (\bz | n)$ captures the prior assumptions about the underlying object and $p_\theta^x(x|\bz)$ defines how it is rendered in the image. 
In general, the inference for \autoref{eq:air} is intractable, so \citep{eslami2016attend} employs amortized variational inference using a sequential algorithm, where an RNN is run for $N$ steps to infer latent representation of one object at a time. 
The variational posterior is then:
\vspace{-5pt}
\begin{align}
\label{eq:air_posterior}
    \q{\bz}{\bx}{\phi} = 
        \q{z_\mathrm{pres}^{n+1} = 0}{\bz^{1:n}, \bx}{\phi} 
        \prod_{i=1}^n 
        \q{\bz^{i}, z_\mathrm{pres}^{i} = 1}{\bz^{1:i-1}, \bx}{\phi},
\end{align}
where $q_\phi$ is a neural network which outputs the parameters of the latent distributions: 
the mean and standard deviation of a Gaussian distribution for $\bz_\mathrm{what}$ and $\bz_\mathrm{where}$ and the probability parameter of the Bernoulli distributed $z_\mathrm{pres}$.

\begin{figure}[t!]
\centering
\begin{minipage}{.5\textwidth}
    \centering
        \includegraphics[width=0.5\textwidth]{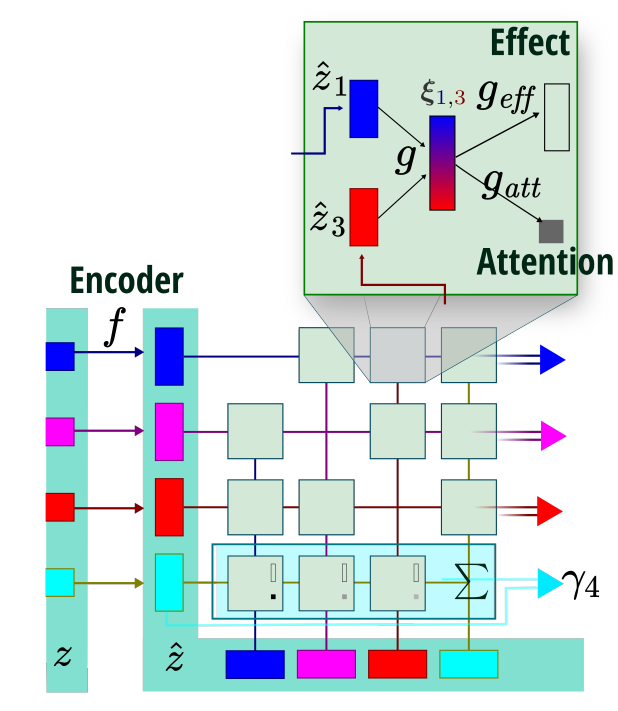}
    \caption[ Interaction Network of R-NEM. ]
    {\small Interaction Network of R-NEM \cite{van2018relational}.} 
    \label{fig:in}
\end{minipage}%
\begin{minipage}{.5\textwidth}
    \centering
        \includegraphics[width=\textwidth]{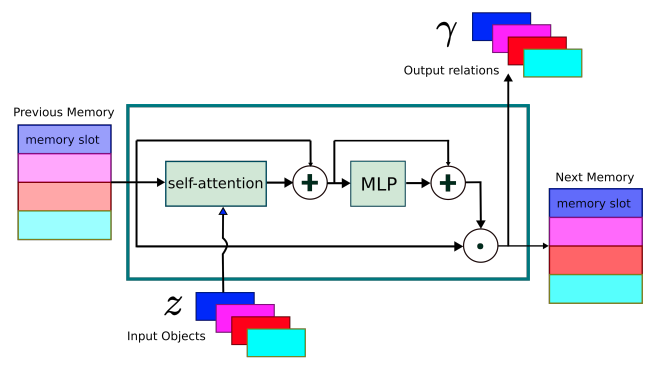}
    \caption[ Relational Memory Core. ]
    {\small Relational Memory Core \cite{santoro2018relational}.} 
    \label{fig:rmc}
\end{minipage}
\end{figure}

\textbf{Relational Sequential Attend, Infer, Repeat (R-SQAIR)} augments SQAIR through a parallel relational mechanism.
SQAIR extends AIR to the sequential setting by leveraging the temporal consistency of objects using a state-space model. It  has two phases: Discovery (DISC) and Propagation (PROP). 
PROP is active from the second frame in the sequence,  propagating or forgetting objects from the previous frame. 
It does so by combining an RNN, which learns temporal dynamics of each object, with the AIR core which iterates over previously propagated objects (explaining away phenomena). 
DISC phase uses the AIR core, conditioned on propagated objects, to discover new appearances of objects. 
For a full description of AIR and SQAIR we refer to previous work \citep{eslami2016attend, kosiorek2018sequential}.

R-SQAIR retains the strengths of its predecessors and improves their relational capabilities. 
More specifically, SQAIR relies on AIR's core RNN to model the relations.
However, an RNN has only a weak relational inductive bias \cite{battaglia2018relational}, as it needs to compute pairwise interactions between objects sequentially, iterating over them in a specific order. 
R-SQAIR, on the other hand, employs networks with strong relational inductive bias which can model arbitrary relations between objects in parallel. 
To construct conceptually simple yet powerful architectures that support combinatorial generalization,
we use the following two methods: \textit{Interaction Network} (IN) \citep{van2018relational} and \textit{Relational Memory Core} (RMC) \citep{santoro2018relational}.


The R-SQAIR generative model is built by extending the PROP module of SQAIR to include relations $\gamma_t = \Gamma(\bz_{t-1})$, where $\Gamma$ is the relational module and $\bz_{t-1}$ are object representations from the previous timestep, defined as follows:
\begin{equation} \label{eq:full_joint}
p(\bx_{1:T},\bz_{1:T}) 
    = \pd(\bz_1^{D_1}) \prod_{t=2}^T \pd(\bzt^{D_t}|\bzt^{P_t})\pp(\bzt^{P_t}|\mathbf{\gamma_t}) p_{\theta}(\bxt|\bzt),
\end{equation}
The \textit{discovery prior} $\pd(\bzt^{D_t}|\bzt^{P_t})$ samples latent variables $\bzt^{D_t}$ for new objects that enter the frame, by conditioning on propagated variables $\bzt^{P_t}$. 
The \textit{propagation prior} $\pp(\bzt^{P_t}|\gamma_t)$ samples latent variables for objects that are propagated from the previous frame and removes those that disappear. 
Both priors are learned during training. 
We recover the original SQAIR model for $\gamma_t = \bz_{t-1}$. 
The inference model is therefore:
\vspace{-5pt}
\begin{equation} \label{eq:full_q}
    \q{\bzTs}{\bxTs}{\phi} 
        = \prod_{t=1}^T \qd{ \bzt^{{D}_t} }{ \bxt, \bzt^{{P}_t} }{ \phi }
        \prod_{i \in \mathcal{O}_{t-1}} \qp{\bzt^i}{\mathbf{\gamma_t}^{i}, h_{t}^{i}}{\phi},
\end{equation}
where $h_{t}^{i}$ are hidden states of the temporal and AIR core RNNs. 
Discovery $\qd_\phi$ is essentially the posterior of AIR. Again, the difference to SQAIR lies in the propagation module $\qp_\phi$, which receives relations $\gamma_t$ as the input. 

\subsubsection*{Relational Module}

\paragraph{Interaction Network}
Our first relational module is the Interaction Network (IN) of R-NEM \cite{van2018relational}, depicted in \autoref{fig:in}, which is closely related to Interaction Networks \cite{battaglia2016interaction, watters2017visual}.
Here, the effect on object $k$ of all other objects $i \neq k$ is computed by the relational module $\gamma_t = \Gamma^{IN} (\bz_{t-1})$, which in the case of IN is defined as follows (for simplicity we drop time indices):
\vspace{-3pt}
\begin{equation}
\bm{\hat{z}}_{k} = f(\bm{z}_{k}), \ \ \bm{\xi}_{k, i} = g([\bm{\hat{z}}_{k};\bm{\hat{z}}_{i}]), \ \ \bm{E}_{k} = \sum_{i \neq k} g_\textit{\hspace{1pt}att}(\bm{\xi}_{k, i}) \cdot g_\textit{\hspace{1pt}eff}(\bm{\xi}_{k, i}), \ \ \gamma_k = [\bz_k; \be_k],
\label{eq:r-nem}
\end{equation}
where $\bz_i = \{\bz_\mathrm{what}^{\mathrm{i}}, \bz_\mathrm{where}^{\mathrm{i}}, z_\mathrm{pres}^{\mathrm{i}}\}$ from the previous time step. First, each object $\bm{z}_{i}$ is transformed using an MLP $f$ to obtain $\bm{\hat{z}}_{i}$, which is equivalent to a  node embedding operation in a graph neural network.
Then each pair $(\bm{\hat{z}}_{k}, \bm{\hat{z}}_{i})$ is processed by another MLP $g$, which corresponds to a  node-to-edge operation by encoding the interaction between object $k$ and object $i$  in the embedding $\bm{\xi}_{k,i}$. 
Note that the computed embedding is directional.
Finally, an edge-to-node operation is performed, where the effect on object $k$ is computed by summing the individual effects $g_\textit{\hspace{1pt}eff}(\bm{\xi}_{k, i})$ of all other objects $i$ on the particular object $k$.
Note that the sum is weighted by an attention coefficient $ g_\textit{\hspace{1pt}att}(\bm{\xi}_{k, i})$, which allows each individual object to consider only particular interactions. 
This technique also yields better combinatorial generalization to a higher number of objects, as it controls the magnitude of the sum.

\paragraph{Relational Memory Core}
We compare the effects modeled by IN to the effects learned by a Relational Memory Core (RMC), $\gamma_t = \Gamma^{RMC} (\bz_{t-1})$. 
RMC (\autoref{fig:rmc}) learns to compartmentalize objects into memory slots, and can keep the state of an object and combine this information with the current object's representation $\bz_t$.
This is achieved by borrowing ideas from memory-augmented networks \citep{sukhbaatar2015end, graves2014neural, graves2016hybrid} and interpreting memory slots as object representations. 
The interactions between objects are then computed by a multi-head self-attention mechanism \citep{vaswani2017attention}
Finally, recurrence for the sequential interactions is introduced, resulting in an architecture that is akin to a 2-dimensional LSTM\citep{hochreiter1997long}, where rows of the memory matrix represent objects.
The model parameters are shared for each object, so the number of memory slots can be changed without affecting the total number of model parameters. 
For a full description, we refer to previous work \citep{santoro2018relational}.
\section{Experiments}

We analyze the physical reasoning capabilities of R-SQAIR on the \textit{bouncing balls} dataset, which consists of video sequences of 64x64 images.
As done in SQAIR experiments, we crop the central 50x50 pixels from the image, such that  a ball can disappear and later re-appear. 
Although visually simple, this dataset contains highly complex physical dynamics and has been previously used for similar studies (R-NEM \citep{van2018relational}). 
The method is trained in SQAIR-like fashion by maximizing the importance-weighted evidence lower-bound IWAE \citep{burda2015importance}, with $5$ particles and the batch size of $32$.
Curriculum learning starts at sequence length 3 which is increased by one every 10000 iterations, up to a maximum length of 10. 
Early stopping is performed when the validation score has not improved for 10 epochs. 

\begin{figure*}[t!]
\centering
\begin{subfigure}[b]{\textwidth}
    \centering
    \includegraphics[width=\textwidth]{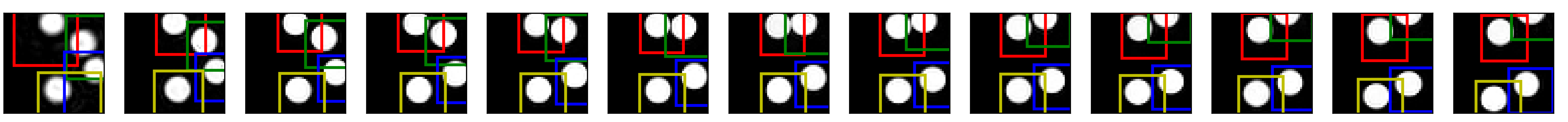}
    \label{fig:bb4_imgs}
\end{subfigure}\\[-3ex]
\begin{subfigure}[b]{\textwidth}
    \centering
    \includegraphics[width=\textwidth]{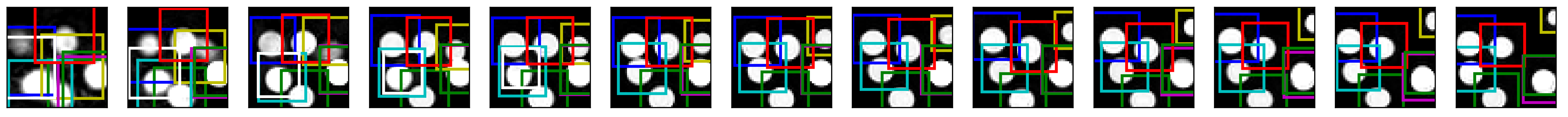}
    \label{fig:bb678_imgs}
\end{subfigure}
\caption[ R-SQAIR trained on sequence of 4 bouncing balls (top row) and evaluated on 6-8 bouncing balls. ]
{\small R-SQAIR trained on sequence of 4 bouncing balls (top rows) and evaluated on 6-8 bouncing balls.} 
\label{fig:bb_imgs}
\end{figure*}

\begin{figure*}
\centering
\begin{subfigure}[b]{0.24\textwidth}
    \centering
    \includegraphics[width=\textwidth]{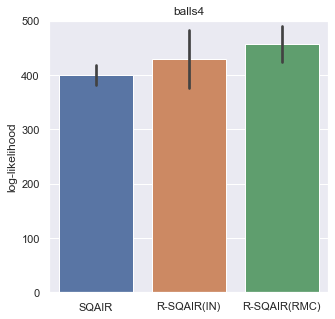}
    \label{fig:bb4_l_ll}
\end{subfigure}
\begin{subfigure}[b]{0.24\textwidth}   
    \centering 
    \includegraphics[width=\textwidth]{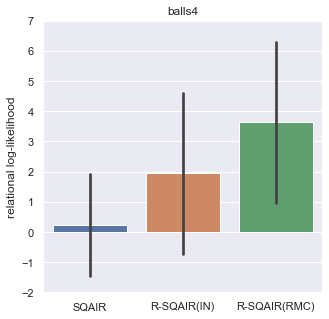}
    \label{fig:bb4_l_r}
\end{subfigure}
\ \ \ 
\begin{subfigure}[b]{0.24\textwidth}  
    \centering 
    \includegraphics[width=\textwidth]{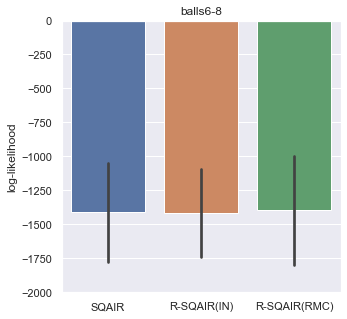}
    \label{fig:bb678_l_ll}
\end{subfigure}
\begin{subfigure}[b]{0.24\textwidth}   
    \centering 
    \includegraphics[width=\textwidth]{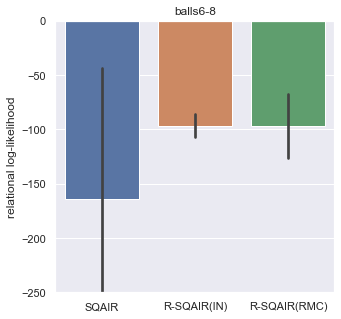}
    \label{fig:bb678_l_r}
\end{subfigure}
\caption[ Log-likelihood and relational log-likelihood of R-SQAIR and SQAIR on bouncing balls task. All models are trained on the balls 4 task.  ]
{\small Log-likelihood and relational log-likelihood of R-SQAIR and SQAIR on the bouncing balls task.} 
\label{fig:bb_quant}
\end{figure*}


Qualitative evaluation of R-SQAIR is present in \autoref{fig:bb_imgs}. 
Each column represents one time step in the video. 
The first row is about the R-SQAIR model trained and evaluated on videos with 4 balls, with object representations highlighted by different color bounding boxes.
In the second row the same model is evaluated on datasets with 6-8 balls.
Note that R-SQAIR disentangles objects already in the first few frames and later only refines the learned representations. At each time step, it computes up to $k=4$ object representations, by considering objects from the previous frame and the learned dynamics.

For all SQAIR hyperparameters we use default values, except for the dimensionality of latent variable $\bz_\mathrm{what}$, which is set to 5 instead of 50. 
This reflects the low visual complexity of individual objects in the scene.
For similar reasons, the embedding dimensionality of IN we use is also set to 5. 
We use a version of the IN module with attention coefficients to compute the weighted sum of the effects.
In total, this adds 9 389 parameters to the 2 726 166 of the default SQAIR implementation.
It also suggests that improved performance is a result of learning a better \textit{propagation prior} instead of just increasing the number of model parameters.

RMC has more hyperparameters to choose from. 
We use self-attention with 4 heads, each of dimensionality 10.
The number of memory slots is 4 and coincides with the total number of sequential attention steps we perform.
Finally, RMC can perform several computations of attention per time step, where each corresponds to one message passing phase.
As we are interested only in collisions, we compute attention only once per time step. 
This results in 98 880 parameters. Comparing the size of the SQAIR model, we obtain a conclusion similar to the one for the case of IN.

Note that the last frames in \autoref{fig:bb_imgs} are sampled from the learned propagation prior.
This enables us to evaluate the role of the relational module, as it is responsible for learning the object dynamics. 
Moreover, as the models are stochastic, we train 5 models for each architecture and sample 5 different last frames.
We compare models in terms of data log-likelihood and relational log-likelihood, which takes into account \textit{only} the objects which are \textit{currently colliding} (ground truth available in the dataset).
The evaluation on the test set with 4 balls shows an increase in average data log-likelihood from 399.5 achieved by SQAIR (0.21 relational) to 429.2 by R-SQAIR(IN) (relational 1.95) and 457.32 by R-SQAIR(RMC) (relational 3.62).
Error bars in \autoref{fig:bb_quant} represent the standard deviation of the stochastic samples from the trained models.

We test generalization of R-SQAIR by evaluating the models trained on sequences with 4 balls on a test set with videos of 6-8 balls.
Both qualitative (\autoref{fig:bb_imgs} bottom row) and quantitative results show that R-SQAIR is capable of generalizing, with increase in relational log-likelihood from -164.1 achieved by SQAIR to -96.7 achieved by R-SQAIR(IN) and -97 achieved by R-SQAIR(RMC).
Larger margins between relational losses of R-SQAIR and SQAIR on the test set with 6-8 balls suggest higher generalization capabilities of R-SQAIR.

\section{Conclusion}

Graph neural networks are promising candidates for
combinatorial generalization, a central theme of AI research \citep{battaglia2018relational, van2019perspective}.  
We show that a sequential attention model can benefit from incorporating an explicit relational module which infers pairwise object interactions in parallel. 
Without retraining, the model generalizes to scenarios with more objects. 
Its learned generative model is potentially useful as part of a world simulator \citep{schmidhuber90sandiego, schmidhuber90diffgenau, ha2018recurrent, watters2019cobra}.

\subsubsection*{Acknowledgments}

We would like to thank Adam R. Kosiorek, Hyunjik Kim, Ingmar Posner and Yee Whye Teh for making the codebase for the SQAIR model \cite{kosiorek2018sequential} publicly available. 
This work was made possible by their commitment to open research practices.
We thank Sjoerd van Steenkiste for helpful comments and fruitful discussions.
This research was supported by the Swiss National Science Foundation grant 407540\_167278 EVAC - Employing Video Analytics for Crisis Management. 
We are grateful to NVIDIA Corporation for  a DGX-1 as part of the Pioneers of AI Research award, and to IBM for donating a ``Minsky'' machine.

\small

\bibliographystyle{abbrv}
\bibliography{biblio}

\end{document}